\documentclass[10pt,twocolumn,letterpaper]{article}

\usepackage[pagenumbers]{wacv} %

\usepackage{graphicx}
\usepackage{amsmath}
\usepackage{amssymb}
\usepackage{booktabs}
\usepackage{multirow}

\usepackage{arydshln}

\usepackage[pagebackref,breaklinks,colorlinks]{hyperref}

\usepackage[capitalize]{cleveref}
\crefname{section}{Sec.}{Secs.}
\Crefname{section}{Section}{Sections}
\Crefname{table}{Table}{Tables}
\crefname{table}{Tab.}{Tabs.}

\begin{document}

\title{D-PoSE: Depth as an Intermediate Representation\\ for 3D Human Pose and Shape Estimation}

\author{Nikolaos Vasilikopoulos\\
Computer Science Department, University of Crete, and\\ Institute of Computer Science, FORTH\\
{\tt\small nvasilik@ics.forth.gr}
\and
Drosakis Drosakis\\
Institute of Computer Science, FORTH\\ 
{\tt\small drosakis@ics.forth.gr}
\and
Antonis Argyros\\
Computer Science Department, University of Crete, and\\ Institute of Computer Science, FORTH\\
{\tt\small argyros@ics.forth.gr}
}
\maketitle

\begin{abstract}
    We present D-PoSE (\textbf{D}epth as an Intermediate Representation for 3D Human \textbf{Po}se and \textbf{S}hape \textbf{E}stimation), a one-stage method that estimates human pose and  SMPL-X shape parameters from a single RGB image. Recent works use larger models with transformer backbones and decoders to improve the accuracy in human pose and shape (HPS) benchmarks. D-PoSE proposes a vision based approach that uses the estimated human depth-maps as an intermediate representation for HPS and leverages training with synthetic data and the ground-truth depth-maps provided with them for depth supervision during training. Although trained on synthetic datasets, D-PoSE achieves state-of-the-art performance on the real-world benchmark datasets, EMDB and 3DPW. Despite its simple lightweight design and the CNN backbone, it outperforms ViT-based models that have a number of parameters that is larger by almost an order of magnitude. D-PoSE code is available at : \href{https://github.com/nvasilik/D-PoSE}{https://github.com/nvasilik/D-PoSE}
\end{abstract}

\section{Introduction}
Vision-based 3D human pose and shape (3D HPS) estimation is an important computer vision research topic with many impactful applications in several application domains. There is already a number of effective solutions for the problem of 2D human body joints estimation from RGB images that are based on neural network architectures~\cite{cao2017realtime,Kreiss_2019_CVPR,sun2019deep}. Therefore, the emphasis has moved to the problems of 3D pose~\cite{zheng20213d,zeng2020srnet} and 3D mesh estimation~\cite{guler2018densepose,lin2021end,kolotouros2019spin,li2022cliff,Kocabas_PARE_2021} for the whole body and  its parts~\cite{xiang2019monocular,black2023bedlam}. 
\begin{figure}[t]
    \centering
    \includegraphics[width=\linewidth]{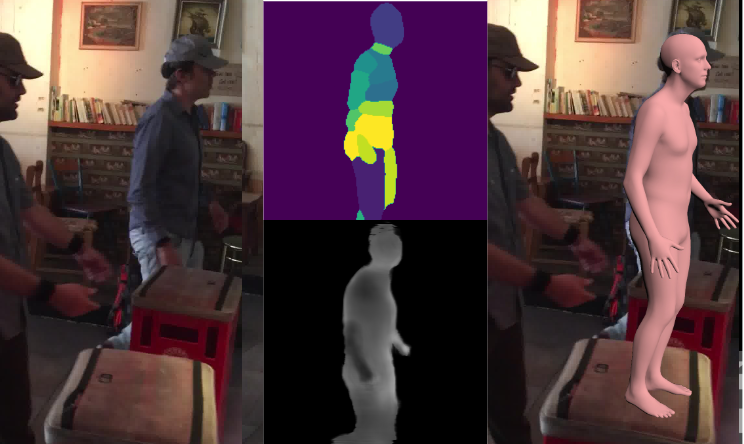}
    \caption{D-Pose, the proposed 3D human Pose and Shape Estimation method receives a single RGB image as input (left), produces intermediate depth and part segmentation representations (middle, bottom and top, respectively) so as to deliver the 3D pose and shape of the imaged person. Despite entailing a small fraction of the parameters of current models, D-PoSE outperforms the current state of the art in 3D pose and shape estimation accuracy in the major relevant datasets (3DPW, EMDB).}
    \label{fig:concept}
\end{figure}

Still, despite the plethora of approaches that have already been proposed, 3D HPS estimation remains a challenging task. Several approaches use video as input~\cite{WeiLin2022mpsnet,kocabas2019vibe}, or depth information provided by RGB-D cameras~\cite{bashirov2021real}.
The recovery of human 3D pose and shape form a single RGB image lacks temporal and depth information and, thus,  has to rely on minimal information to solve very challenging 2D to 3D ambiguities. Therefore, this is the most challenging of all settings. At the same time, this is the most general setting that makes the least amount of assumptions regarding the input of the estimation problem. Therefore, a robust and accurate solution given the minimal input of a single image frame is very general and can be very impactful in a number of application domains.

In this work, we focus on this challenging version of the 3D HPS estimation problem where 3D human pose and shape have to be recovered on the basis of a single RGB frame, only (see Fig.~\ref{fig:concept}). To address this problem, we propose D-PoSE, a method that leverages ground-truth depth maps from recent synthetic datasets and learns to predict human depth maps that incorporates them in the prediction procedure for more accurate 3D HPS estimation.
Specifically, D-PoSE uses synthetic RGB data as input, together with the associated depth maps which are only used for supervision during training and not as input at run-time. In our work human depth maps serve as an intermediate representation, together with an  estimated human body parts segmentation.

The training of D-PoSE capitalizes on the the availability of synthetic data. With the introduction of the recent BEDLAM synthetic dataset~\cite{black2023bedlam}, models are able to train only with synthetic data and outperform the accuracy of training with real-world data. BEDLAM provides accurate synthetic depth maps with ground-truth 3D keypoints and SMPL-X~\cite{SMPL-X:2019}
parameters. Although there is a domain gap between synthetic and real-world depth, the proposed model generalizes well in real-world datasets.

Current state-of-the-art methods~\cite{dwivedi2024tokenhmr,hmrKanazawa17} use ViT~\cite{dosovitskiy2020vit} backbones.
While those backbones benefit from large datasets, they increase dramatically the model size. Therefore, those methods need long training times and multiple flagship GPUs. One of the goals of our work is to provide a lightweight vision-based solution to the 3D HPS estimation problem that has state-of-the-art performance without the need of extra training dataset(s) and does not employ oversized models. 

We demonstrate that the use of depth information as an intermediate representation together with part segmentation on a  simple CNN backbone suffices to deliver state of the art results in terms of both accuracy and model size. Specifically, we performed several experiments on the challenging 3DPW~\cite{vonMarcard2018} and EMDB~\cite{kaufmann2023emdb} datasets.
The experimental results demonstrate improvements of 3.0mm in PA-MPJPE, 3.1mm in MPJPE and 3.6mm in MVE when compared with BEDLAM-CLIFF~\cite{black2023bedlam} in the challenging 3DPW dataset.
When compared with the state-of-the-art method TokenHMR~\cite{dwivedi2024tokenhmr} which employs a ViT backbone, our method reduces error by 0.4mm in PA-MPJPE, 2.7mm in MPJPE and 4.3mm in MVE. At the same time, our model has 83.8\% less parameters than TokenHMR.

In summary, the main contributions of this work are the following: 
\begin{itemize}
    \item We propose D-PoSE, a novel method to the problem of 3D HPS estimation from a single RGB frame. D-PoSE uses depth information from synthetic data as an intermediate representation and generalizes well to to real-world data.
    \item We demonstrate that D-PoSE achieves state-of-the-art accuracy in Mean Vertices Error (MVE) and Mean per Joint Position Error (MPJPE) in standard HPS benchmarks.
    \item We also demonstrate that D-PoSE entails significantly less trainable model parameters, specifically 83.8\% less parameters compared to the current state-of-the-art method.
\end{itemize}

\section{Related Work}
The 3D HPS estimation problem has been approached in several ways, including optimization-based techniques (usually by fitting a mesh to 2D keypoints), learning-based techniques (where a model is trained to predict a 3D mesh).  
We also review various intermediate representations that have been employed as well as training datasets that are relevant to our work.
D-PoSE is a one-stage, learning-based method which takes a single RGB image as input uses intermediate representations before estimating the 3D mesh.

\vspace*{0.1cm}\noindent\textbf{Optimization-based methods}:
Optimization approaches use 2D image cues to fit a parametric model. Bogo \etal~\cite{bogo2016keep} proposed Simplify, which optimizes the 3D shape and pose of the SMPL~\cite{SMPL:2015} human model
using 2D keypoints. Omran \etal~\cite{omran2018} proposed the use of silhouettes to handle perspective ambiguities. Lassner \etal~\cite{Lassner:UP:2017} used part segmentation to improve the body shape and pose estimation. Optimization approaches require less data but are prone to 2D-3D ambiguities.

\begin{figure*}[t]
    \centering
    \includegraphics[width=\linewidth]{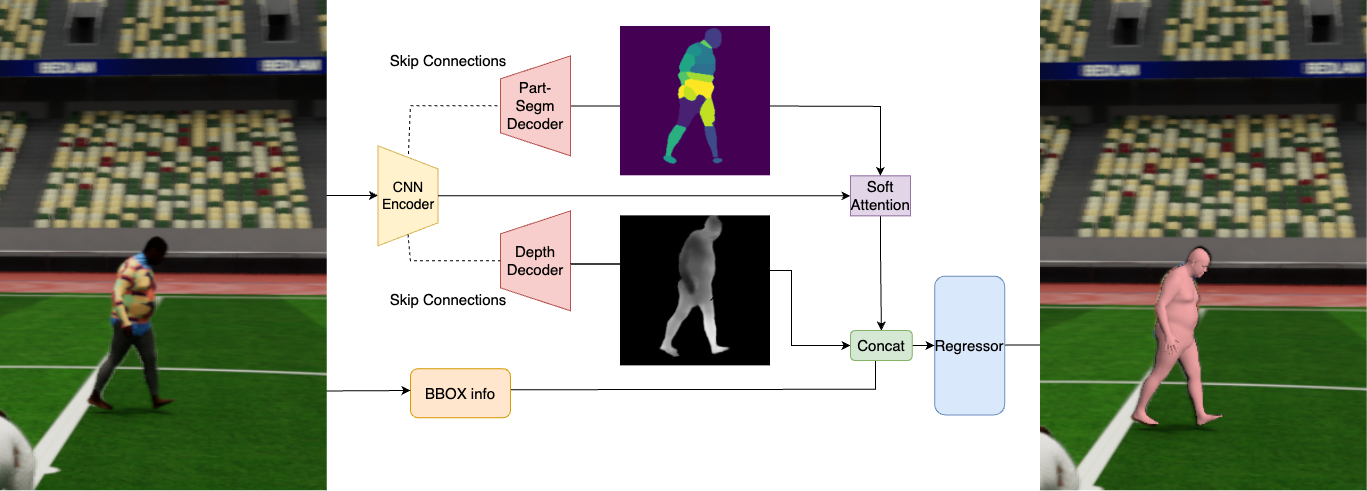}
    \caption{The architecture of D-PoSE. Given an input image, features are extracted using a CNN. With these feature maps a human depth map and a part-segmentation map are estimated. The original features pass through a soft-attention mechanism which uses part-segmentation maps. The final features are concatenated with the bounding-box information and the depth features and are given as input to the regressor which estimates the 3D human pose and shape.}
    \label{fig:enter-label}
\end{figure*}

\vspace*{0.1cm}\noindent\textbf{Learning-based methods}:
Learning based approaches estimate directly model parameters~\cite{hmrKanazawa17, dwivedi2021, Kocabas_PARE_2021, sun2021, Sun2021PuttingPI, choi2022, li2022cliff, poco}. A model-free representation can be estimated such as vertices~\cite{KolotourosMeshRegression, MeshGraphormer, Srndi2020MeTRAbsMT} or implicit shape~\cite{COAP, PIFuHD, ECON}. Li \etal~\cite{Li2020HybrIKAH} proposed a novel hybrid inverse kinematics solution (HybrIK) which computes the 3D joint positions of a human body by combining an analytical solution and a neural network regression. Pose priors can also be employed, imposing constraints on the human pose and shape in order to reduce invalid estimations. These could include joint limits~\cite{Akhter2015PoseConditioned} where they would prune invalid human poses, Gaussian Mixture Models~\cite{bogo2016keep}, Generative Adversarial Networks~\cite{Georgakis2020HierarchicalKH, hmrKanazawa17}, VAEs~\cite{Pavlakos2019ExpressiveBC} and normalizing flows~\cite{prohmr} that can be used as knowledge priors in the training process. Kolotouros \etal~\cite{kolotouros2019spin} proposed SPIN which improves the pose estimation accuracy by fitting the body model to 2D keypoints in the training loop. CLIFF~\cite{li2022cliff} provides the neural network with information about the bounding box coordinates containing the human in the image, gaining a noticeable accuracy improvement. These methods can have less ambiguities but rely on additional data for robust training.

\vspace*{0.1cm}\noindent\textbf{Intermediate representations}:
Intermediate representations could allow for more training data to be injected in the training process. HoloPose~\cite{guler2019holopose} aligns initial 3D part based model prediction with the 2D keypoints, 3D keypoints and DensePose~\cite{guler2018densepose}.
More recently, Kocabas~\etal~\cite{Kocabas_PARE_2021} proposed PARE, a model that uses a part-segmentation branch together with an attention mechanism in order to achieve an occlusion-robust method. Our part-segmentation branch is highly inspired by PARE but deviates considerably from it with respect to specific choices in its architecture.
Zhu~\etal~\cite{zhu2019detailed,zhu2022detailed} (HMD) argued that by utilizing per-pixel shading information and depth, it is possible to refine the shape and produce a detailed 3D mesh with deformations. Although HMD suggests the use of depth, our method directly uses it in pose and shape estimation process and does not deform the SMPL-X mesh based on the depth.

\vspace*{0.1cm}\noindent\textbf{Depth estimation}:
Varol~\etal~\cite{varol2017learning} suggested to train a CNN with synthetic data and use them to to predict human depth maps and human part-segmentation maps. However neither the human depth nor the segmentation map was used for pose or shape estimation.
Zhou~\etal found that DIFFNet~\cite{zhou_diffnet} with HRNet as encoder and a UNET-like depth decoder is effective in standard depth estimation datasets.

\vspace*{0.1cm}\noindent\textbf{Synthetic data}:
AGORA~\cite{Patel:CVPR:2021} provides SMPL-X ground truth data and synthetic images of clothed humans generated from static commercial scans. The inclusion of AGORA in training datasets enhances the accuracy of 3D HPS estimation methods. Additionally, AGORA serves as a benchmark for evaluating 3D HPS estimation approaches.

Black \etal~\cite{black2023bedlam} proposed a new synthetic dataset with ground truth SMPL-X data, realistic human images and depth maps named BEDLAM. Training HMR ~\cite{hmrKanazawa17} and CLIFF~\cite{li2022cliff} using the BEDLAM dataset proves itself enough to achieve state-of-the-art performance. The same work also suggests the use of vertices loss.

\vspace*{0.2cm}\noindent\textbf{Vision transformers backbone}:
HMR2.0~\cite{goel2023humans} uses ViT backbone to encode the image and a transformer based decoder to predict the 3D mesh.
TokenHMR~\cite{dwivedi2024tokenhmr}, using the same backbone with HMR2.0, reformulates the problem of HPS by  tokenizing the pose tokens in the encoder and letting the decoder reconstruct the original pose. 

\vspace*{0.2cm}\noindent\textbf{The proposed D-PoSE approach}: The proposed D-PoSE is a one-stage method that takes a single RGB image as input and estimates two intermediate representations: (1)~human depth and (2)~part-segmentation of the human. Using these representations, along with the original CNN features, it regresses the 3D human pose and shape. D-PoSE does not use vision transformers as backbone but a CNN.

\section{Methodology}
An overview of the architecture of D-PoSE is provided in Fig.~\ref{fig:enter-label}. Given an input image, features are extracted using a CNN. With these feature maps a human depth map and a part-segmentation map are estimated. The CNN features pass through a soft-attention mechanism which uses the part-segmentation maps. The final features are concatenated with the bounding-box information and the estimated human depth map
and are given as input to the regressor which estimates the 3D human pose and shape. Below, we provide further details on each and every of the aforementioned modules and representations. 

\subsection{CNN Encoder}
D-PoSE uses the High-Resolution Network (HRNet-W48)~\cite{cheng2020bottom,SunXLW19,wang2019deep} as the convolutional neural network (CNN) encoder. The HRNet-W48 is selected for its ability to produce spatially precise feature maps at multiple resolutions. Specifically, given a single RGB image input of dimensions $256 \times 256$, the encoder generates a set of feature maps at four distinct resolutions: $\mathbf{F}_1 \in \mathbb{R}^{384 \times 7 \times 7}$, $\mathbf{F}_2 \in \mathbb{R}^{192 \times 14 \times 14}$, $\mathbf{F}_3 \in \mathbb{R}^{96 \times 28 \times 28}$ and $\mathbf{F}_4 \in \mathbb{R}^{48 \times 56 \times 56}$. These feature maps are utilized in skip-connections with the decoder layers to enhance the reconstruction process. Additionally, the encoder outputs an up-sampled feature vector 
$\mathbf{F}_\text{down} \in \mathbb{R}^{720 \times 56 \times 56}$. HRNet-W48 has previously been used in BEDLAM-CLIFF~\cite{black2023bedlam}
and BEDLAM-HMR~\cite{black2023bedlam}
and is a standard choice for CNN backbones used in pose estimation tasks.

\subsection{Human Models}
To predict the 3D human mesh, we utilize the SMPL-X~\cite{SMPL-X:2019} body model, which consists of $N = 10{,}475$ vertices and $K = 54$ joints, including those for the neck, jaw, eyeballs and fingers. The SMPL-X model is represented by the function $M(\theta, \beta, \psi)$, where $\theta$ denotes pose parameters, $\beta$ captures shape parameters, and $\psi$ represents facial expression parameters.

\begin{figure}[t]
    \centering
    \includegraphics[width=\linewidth]{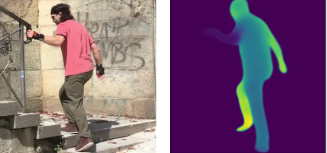}
    \caption{Left: Image sampled from 3DPW, Right: human depth-map estimated by our method.}
    \label{fig:depth}
\end{figure}

For evaluations on the 3DPW~\cite{vonMarcard2018} and EMDB~\cite{kaufmann2023emdb} datasets, we transform our predicted SMPL-X~\cite{SMPL-X:2019} meshes into SMPL~\cite{SMPL:2015} format by applying a vertex mapping matrix $D \in \mathbb{R}^{10475 \times 6890}$. This conversion is used exclusively for assessing body pose and shape. Similarly, we convert the ground truth SMPL-X vertices to SMPL format using $D$ after neutralizing the hand and face poses. To calculate joint errors, we extract $22$ joints from the vertices using the SMPL joint regressor.

For both SMPL and SMPL-X we use the gender neutral models.

\subsection{Loss function}
The loss function $L$ consists of three parts, depth loss, segmentation loss and 3D human loss:
\begin{equation}
    L = L_{depth} + L_{segm} + L_{human}.
\end{equation}

\vspace*{0.1cm}\noindent\textbf{Depth loss}: For the depth term loss, background is ignored and a combination of L1 loss and structural similarity index measure is used:
\begin{multline}
    L_{depth}= \lambda_{1}|depth_{gt} - depth_{pred}| + \\ \lambda_{2}(1 - SSIM(depth_{gt},depth_{pred})).
\end{multline}

\vspace*{0.1cm}\noindent\textbf{Part segmentation loss}: To produce accurate part-segmentation, we use cross entropy loss between the predicted and ground-truth SMPL part-segmentations:
\begin{equation}
    L_{segm} = \lambda_{3} CrossEntropy(gt,pred).
\end{equation}

\begin{figure}[t]
    \centering
    \includegraphics[width=\linewidth]{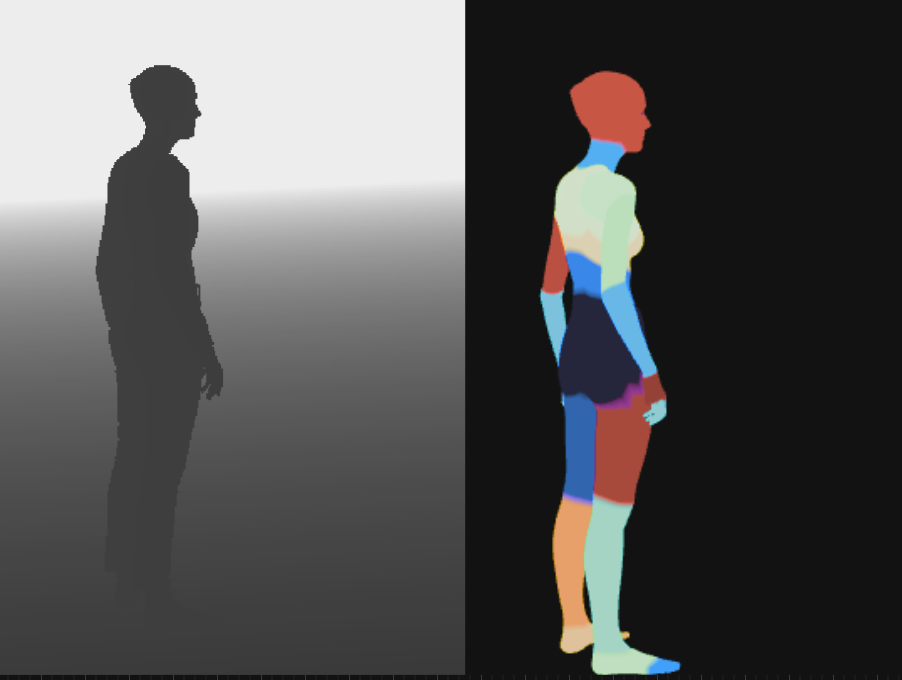}
    \caption{Left: Ground-truth depth-pap visualized in grayscale (BEDLAM dataset). Right: Ground-Truth SMPL-X Mesh after rendering with part-segmentation (BEDLAM dataset).}
    \label{fig:bedlam}
\end{figure}

\vspace*{0.1cm}\noindent\textbf{3D human loss}: For 3D human prediction, as proposed by BEDLAM-CLIFF~\cite{black2023bedlam}, we use two MSE SMPL-X losses one for SMPL-X pose $\theta$ parameters and one for SMPL-X shape $\beta$ parameters, a 3D Joints MSE loss between the ground-truth and estimated 3D Joints,  and the newly proposed 3D vertices loss which is a L1 loss between the ground-truth SMPL-X vertices and the estimated SMPL-X vertices:
\begin{equation*}
    L_{SMPL_{pose}}= ||\hat{\theta} - \theta||,
\end{equation*}
\begin{equation*}
    L_{SMPL_{shape}}= ||\hat{\beta} - \beta||,
\end{equation*}
\begin{equation*}
    L_{J3D}= ||\hat{J}_{3D} - J_{3D}||,
\end{equation*}
\begin{equation*}
    L_{J2D}= ||\hat{J}_{2D}  - J_{2D}||,
\end{equation*}
\begin{equation*}
    L_{V3D}= |\hat{V} - V|.
\end{equation*}
Given the above, the 3D human loss is defined as:
\begin{multline}
       L_{human}= \lambda_{4}L_{SMPLpose} + \lambda_{5}L_{SMPLshape} \\ +\lambda_{6}L_{J3D} +\lambda_{7}L_{V3D} + \lambda_{8}L_{J2D}.
\end{multline}

\subsection{Depth}
Depth is estimated by the depth decoder.
The depth decoder takes as input  the features extracted by the HRNet backbone and uses skip connections from the previous stages to capture hierarchical features [$\mathbf{F}_0 \in \mathbb{R}^{384 \times 7 \times 7}$, $\mathbf{F}_1 \in \mathbb{R}^{192 \times 14 \times 14}$, $\mathbf{F}_2 \in \mathbb{R}^{96 \times 28 \times 28}$ and $\mathbf{F}_3 \in \mathbb{R}^{48 \times 56 \times 56}$]. 
The depth decoder has a U-Net~\cite{ronneberger2015u} like structure. However it is lighter and not symmetrical to the encoder path. The output of the decoder is the relative depth of the human ignoring the background which is represented with zero values. The BEDLAM dataset provides ground truth depth maps stored as $32$-bit float in Unreal coordinate system units. From the depth maps we remove the background using the background mask provided and keep only the values of the human body.  Then we set background values to zero and normalize the rest of the values in the range $[0.1, 1.0]$. A sample depth output is visualized in Fig.~\ref{fig:depth}.

\subsection{Part Segmentation}
Althught the concept of using part segmentation is similar to that of PARE~\cite{Kocabas_PARE_2021}
, the architecture of the part segmentation decoder is similar to that of the depth decoder. The only difference is the last layer which outputs $23$ channels and the body model. While PARE uses SMPL model, we use SMPL-X since we train with the AGORA and BEDLAM datasets.

Since BEDLAM and AGORA provide ground-truth SMPL-X parameters, during training we use these parameters to generate SMPL-X mesh. From the generated ground-truth mesh, we map each vertex to the human joint that it belongs. We end up with $22$ different body parts (PARE that uses SMPL has $24$), each assigned with a different value (see Fig.~\ref{fig:bedlam}). The background is also assigned to the value of zero. Finally, we render the part-segmented SMPL-X mesh and use it to supervise the part-segmentation.

In contrast to PARE, part-segmentation remains supervised throughout the entire training process and is not disabled at any point.

\subsection{Soft-Attention}
The soft-attention mechanism employed in our work is similar to that used in PARE. It takes as input a tensor $\mathbf{F_{\text{upsampled}}} \in \mathbb{R}^{720 \times 56 \times 56}$ which is the CNN features . Additionally, it processes the part-segmentation images $\mathbf{S} \in \mathbb{R}^{23 \times 56 \times 56}$. The part-segmentation tensor $\mathbf{S}$ is passed through a softmax operation over the spatial dimensions while ignoring the first segmentation-map which is attributed to background, producing normalized attention maps $\mathbf{\sigma(S}) \in\mathbb{R}^{22\times(56\times56)}$.

In order to produce the attention-weighted features, we first reshape the feature vector $\mathbf{F}_{\text{reshaped}} \in \mathbb{R}^{720 \times (56 \times 56)}$. The attention-weighted features are obtained by:
\begin{equation}
    A = \sigma(S) \cdot F_{\text{reshaped}}^{T}.
\end{equation}
For the shape of tensor $A$ it holds that  $A \in \mathbb{R}^{22 \times 720}$.%

\subsection{Bounding Box}
As proposed by CLIFF, we supervise the 2D reprojection
loss in the original full-frame image instead of the cropped image.  Specifically, 
\begin{equation}
	J_{2D}^{full} = \mathrm{\Pi} J_{3D}^{full}
			      = \mathrm{\Pi} (J_{3D} +  \mathbf{t}^{full}),
\end{equation}
where $\mathbf{t}^{full}$ represents the translation relative to the optical center of the original image.
Also, we concatenate the bounding-box center and scale with the features produced by the attention mechanism. As a result, the estimated global orientation is improved.

\subsection{Decoders Architecture}

The architecture of the decoders employed in our model consists of a series of upsampling and refinement modules that progressively refine the feature maps obtained from the backbone network. The model is designed to produce a depth map or part segmentation map from these feature maps.

\vspace*{0.1cm}\noindent\textbf{Input:}
Let $\mathbf{F}_i \in \mathbb{R}^{B \times C_i \times H_i \times W_i}$ represent the feature maps from the backbone network at resolution level $i$, where $B$ is the batch size, $C_i$ is the number of channels, and $H_i \times W_i$ are the spatial dimensions. We denote these feature maps as $\{\mathbf{F}_0, \mathbf{F}_1, \mathbf{F}_2, \mathbf{F}_3\}$ for increasing resolution levels.

\begin{table*}[t]
\centering
\begin{tabular}{ll|l|cccccc}
\cline{2-9}
 & \textbf{Training} & \multicolumn{1}{c|}{\multirow{2}{*}{\textbf{Method}}} & \multicolumn{3}{c|}{\textbf{EMDB~\cite{kaufmann2023emdb}}} & \multicolumn{3}{c}{\textbf{3DPW~\cite{kaufmann2023emdb}}} \\ \cline{2-2} \cline{4-9} 
 & \textbf{\textbf{Datasets}} & \multicolumn{1}{c|}{} & \textbf{MVE} & \textbf{MPJPE} & \multicolumn{1}{c|}{\textbf{PA-MPJPE}} & \textbf{MVE} & \textbf{MPJPE} & \textbf{PA-MPJPE} \\ \cline{2-9} 
\multirow{5}{*}{\rotatebox{90}{\textbf{HRNet}}} & SD & PARE & - & - & \multicolumn{1}{c|}{-} & 97.9 & 82.0 & 50.9 \\
 & SD & CLIFF & - & - & \multicolumn{1}{c|}{-} & 87.6 & 73.9 & 46.4 \\
 & BL & BEDLAM-HMR & - & - & \multicolumn{1}{c|}{-} & 93.1 & 79.0 & 47.6 \\
 & BL & BEDLAM-CLIFF & 113.2 & 97.1 & \multicolumn{1}{c|}{61.3} & 85.0 & 72.0 & 46.6 \\ \cdashline{2-9}
 & BL & \textbf{D-PoSE (Ours)} & \textbf{99.0} & \textbf{85.5} &  \textbf{53.2} & \textbf{81.4} & \textbf{68.9} & \textbf{43.6} \\ \cline{2-9} 
\multirow{2}{*}{\rotatebox{90}{\textbf{ViT}}} & BL & HMR2.0 & 106.6 & 90.7 & \multicolumn{1}{c|}{51.3} & 88.4 & 72.2 & 45.1 \\
 & BL & TokenHMR & 106.2 & 89.6 & \multicolumn{1}{c|}{\textbf{49.8}} & 85.7 & 71.6 & 44.0 \\ \cdashline{2-9} 
\rotatebox{90}{\parbox{0.7cm}{\textbf{HR}\\\textbf{Net}}} & BL & \textbf{D-PoSE (Ours)} & \textbf{99.0} & \textbf{85.5} & \multicolumn{1}{c|}{53.2} & \textbf{81.4} & \textbf{68.9} & \textbf{43.6} \\ \cline{2-9} 
\end{tabular}
\caption{HPS errors on the EMDB and 3DPW datasets. SD represents standar realistic datasets and BL represents training only with synthetic datasets BEDLAM and AGORA. See text.}
\label{tab:comp-sota}
\end{table*}

\vspace*{0.1cm}\noindent\textbf{Up-sampling Modules:}
The upsampling process is performed through a series of upsampling modules. Each module $U_i$ upsamples the feature maps from resolution level $i+1$ to resolution level $i$ using a bilinear interpolation with scale factor equal to $2$ followed by a $1 \times 1$ convolution. Specifically, 
\begin{equation}
\mathbf{F}_i^{\text{up}} = U_i(\mathbf{F}_{i+1}^{\text{up}}) = \text{ReLU}\left(\text{BN}\left(\text{Conv}_{1\times1}\left(\mathbf{F}_{i+1}^{\text{up}}\right)\right)\right),
\end{equation}
where $\mathbf{F}_i^{\text{up}} \in \mathbb{R}^{B \times C_i \times H_i \times W_i}$ is the upsampled feature map at level $i$, and $U_i$ denotes the upsampling operation at level $i$. The upsampled feature map is concatenated with the corresponding feature map from the backbone network. Therefore, 
\begin{equation}
\mathbf{F}_i^{\text{cat}} = \text{Concat}\left(\mathbf{F}_i^{\text{up}}, \mathbf{F}_i\right),
\end{equation}
where $\mathbf{F}_i^{\text{cat}} \in \mathbb{R}^{B \times (C_i + C_i) \times H_i \times W_i}$.

\vspace*{0.1cm}\noindent\textbf{Fusion and Refinement:}
The concatenated feature maps $\mathbf{F}_i^{\text{cat}}$ are passed through a fusion and refinement module $R_i$, which consists of 4 residual blocks:
\begin{equation}
\mathbf{F}_i^{\text{ref}} = R_i(\mathbf{F}_i^{\text{cat}}),
\end{equation}
where $\mathbf{F}_i^{\text{ref}} \in \mathbb{R}^{B \times C_{i-1} \times H_i \times W_i}$ and $R_i$ denotes the refinement operation at level $i$.

\vspace*{0.1cm}\noindent\textbf{Final Layers:}
The final output map (either a depth map or part-segmentation map) is generated by a series of convolutions, ReLU activation functions and batch normalization layers applied to the output of the lowest resolution refinement module. In notation, 
\begin{equation}
\mathbf{O} = \text{Conv}_{1\times1}\left(\text{ReLU}\left(\text{BN}\left(\text{Conv}_{3\times3}(\mathbf{F}_0^{\text{ref}})\right)\right)\right),
\end{equation}
where $\mathbf{O} \in \mathbb{R}^{B \times C_{\text{out}} \times H_0 \times W_0}$ is the final output, and $C_{\text{out}} = 1$ for depth maps or $C_{\text{out}} = 23$ for part-segmentation maps.

\vspace*{0.1cm}\noindent\textbf{Output:}
The final output consists of a depth map $\mathbf{O}_{\text{depth}} \in \mathbb{R}^{B \times 1 \times H_0 \times W_0}$ or a part-segmentation map $\mathbf{O}_{\text{psegm}} \in \mathbb{R}^{B \times 23 \times H_0 \times W_0}$.

\subsection{Regressor}
To regress the SMPL-X pose parameters we use a regressor with the same MultiLinear layer that ReFit~\cite{wang23refit}
proposes. 
The forward pass is efficiently computed using Einstein summation notation, and bias terms are added per head. The 22 heads representing each joint compute the body pose parameters in parallel.

The three camera parameters and the eleven shape parameters are computed by simple linear layers followed by ReLU activation functions.

Our model demonstrates faster convergence and training speeds with this regressor compared to the PARE~\cite{Kocabas_PARE_2021}
and CLIFF~\cite{li2022cliff}
regressors.

\begin{table}[t]
\centering
\begin{tabular}{l|r}
\hline
\textbf{Method}   & \textbf{Number of Parameters} \\ \hline
HMR2.0   &          672.0 Million         \\ 
TokenHMR &        681.0 Million         \\ 
\textbf{D-PoSE (Ours)}     &            \textbf{81.2 Million}  \\ \hline
\end{tabular}
\caption{Number of parameters of each model.} 
\label{tab:comp_params}
\end{table}

\section{Experiments}
\subsection{Datasets}

D-PoSE is trained solely on synthetic data. 
BEDLAM~\cite{black2023bedlam}
is used subsampled at 6 frames per second as the proposed method BEDLAM-CLIFF. We use the ground truth training data, including the provided depth maps. Also, BEDLAM provides masks for the background which we use to remove it from the depth maps.

AGORA~\cite{Patel:CVPR:2021}
is the the second synthetic dataset we use for training. AGORA is used to supervise the segmentation and 3D human loss but not the depth since it lacks ground truth depth maps.

Our method is evaluated on the 3DPW~\cite{vonMarcard2018}
and EMDB~\cite{kaufmann2023emdb}
datasets. Both of them contain real images of humans in the wild. Since both datasets have ground truth SMPL data, we are able to calculate Mean Vertices Error(MVE) on both datasets to capture the accuracy of the estimated human shapes.

We also use the RICH~\cite{Huang:CVPR:2022}
dataset for obtaining qualitative results and for the ablation study~\ref{tab:abl_depth}. RICH differs from the other datasets by including humans interacting with objects and their environment in both indoor and outdoor scenes.

\begin{figure*}[t]
    \centering
    \includegraphics[width=\linewidth]{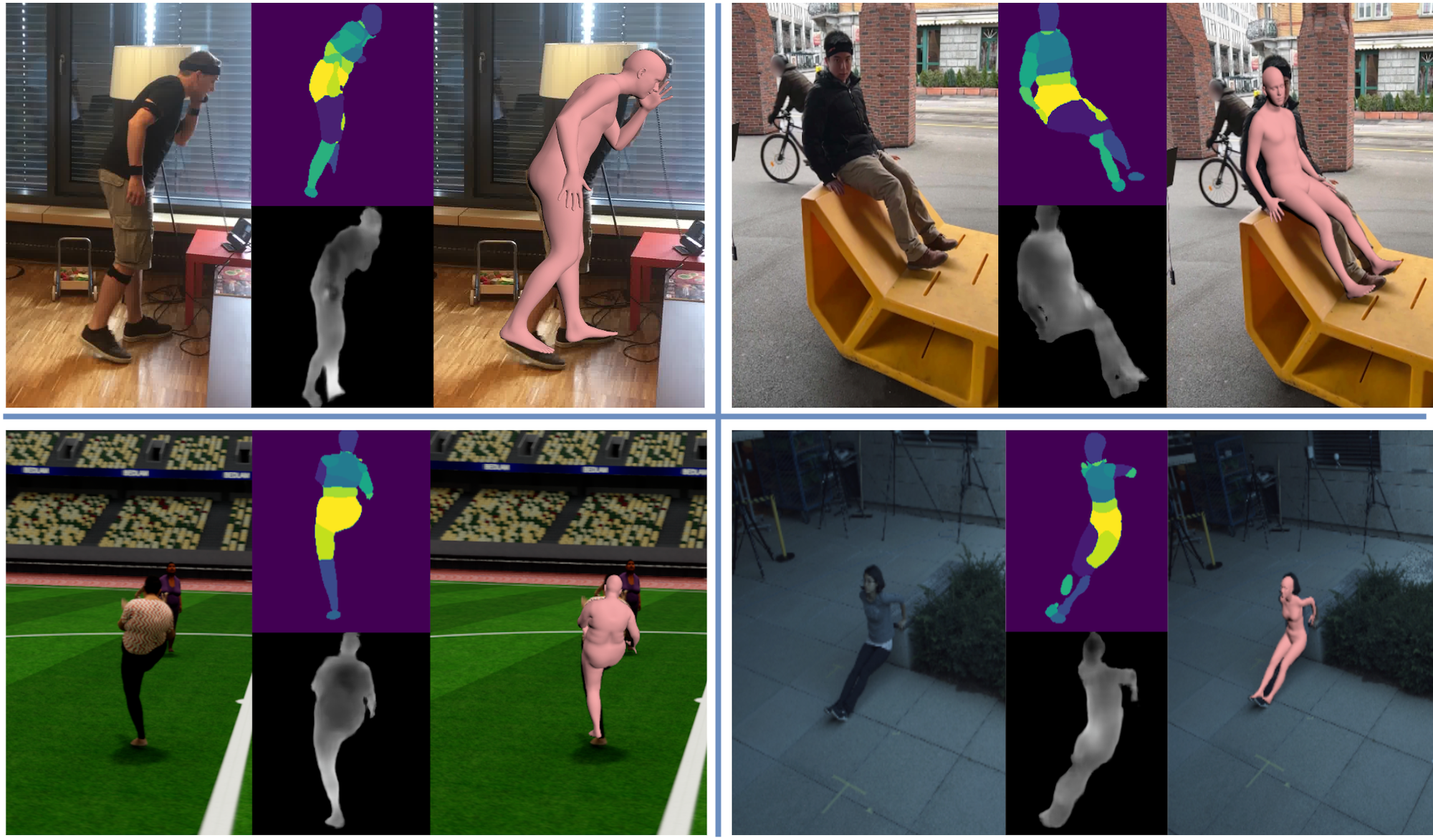}
    \caption{Each image block represents: the input image (left); the part-segmentation estimation as an intermediate representation (middle-top); the human depth map as an intermediate representation (middle-bottom); the 3D HPS estimation of our method (right). The figure illustrates results from the 3DPW dataset (top left block) the EMDB test set (top right), synthetic image sampled from the BEDLAM validation set (bottom left) and from the RICH dataset (bottom right).}
    \label{fig:qualitative2}
\end{figure*}

\subsection{Training}
We train our model using PyTorch in one stage. Training requires $200K$ iterations with batch size of $64$. The optimizer used is Adam with a learning rate of $1e-5$ and zero weight decay. For numerical stability we use gradient clipping with value $1.5$.

A single NVidia RTX-A100 GPU is used for all the experiments. Training in our system requires 3 days.

Random augmentations are applied to the RGB images using Albumentations~\cite{info11020125} similarly with BEDLAM-CLIFF. Those augmentations include 
random cropping, down-scaling, compressing the image, random rain and snow noise, multiplicative noise, motion blur, blurring, random occlusions, CLAHE and equalization, random changes to brightness and contrast,  hue saturation, random gamma and posterization.

We use HRNet-W48 as the CNN backbone to extract features from the RGB image in four resolutions. The size of the input RGB image is $224 \times 224$. HRNet-W48 is initialized with weights pretrained on COCO~\cite{lin2014microsoft}. The Neural 3D Mesh Renderer~\cite{kato2018renderer} is used to render the part-segmented SMPL mesh during training.

For fair comparison with BEDLAM-CLIFF we use the $80\%$ of BEDLAM and AGORA for training.

The coefficients of the loss functions for the experiments are: %
$\lambda_{1}=0.1$, $\lambda_{2}=0.02$, $\lambda_{3}=0.1$, $\lambda_{4}=10$, $\lambda_{5}=0.01$, $\lambda_{6}=50$, $\lambda_{7}=10$ and $\lambda_{8}=50$.

\subsection{Evaluation Metrics}
For the quantitative evaluation of D-PoSE we use the following well-established evaluation metrics: 

\vspace*{0.1cm}\noindent\textbf{Mean Per Joint Position Error (MPJPE)}: MPJPE aligns the predicted and ground-truth 3D joints at the pelvis and measures the resulting distances, providing a comprehensive evaluation of pose and shape, including global rotations. 

\vspace*{0.1cm}\noindent\textbf{Procrustes-Aligned MPJPE (PA-MPJPE)}: PA-MPJPE applies Procrustes alignment before calculating MPJPE, focusing on articulated pose accuracy by removing scale and rotation discrepancies. 

\vspace*{0.1cm}\noindent\textbf{Mean Vertex Error (MVE)}: MVE also considers pelvis alignment of the predicted and ground-truth 3D joints but evaluates the distances between vertices on the human mesh surface.

\begin{figure*}[t]
    \centering
    \includegraphics[width=0.99\linewidth]{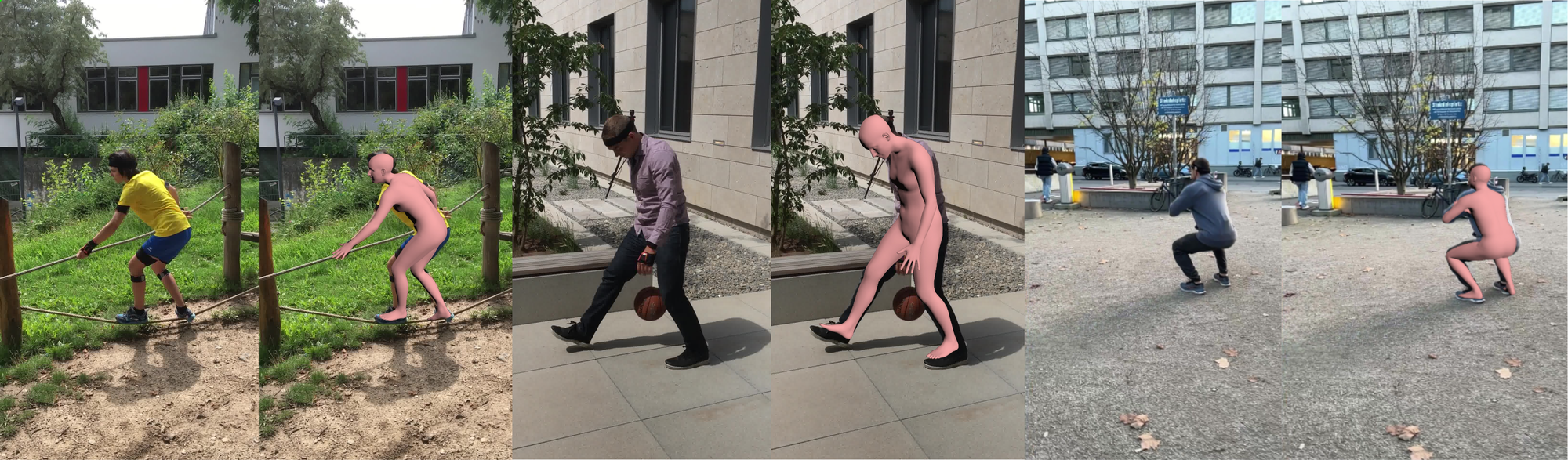}
    \caption{%
    Further qualitative  results sampled from the challenging 3DPW~\cite{vonMarcard2018} and EMDB~\cite{kaufmann2023emdb} datasets.
    }
    \label{fig:header}
\end{figure*}

\subsection{Quantitative Results}
In Table \ref{tab:comp-sota} we compare our method with the current state of the art methods. In order to evaluate our method we convert 3DPW and EMDB SMPL meshed to SMPL-X.
In both 3DPW and EMDB we report 
Mean Vertex Error (MVE) using the vertices obtained from the SMPL mesh, Mean Per Joint Position Error (MPJPE) of the human 3D joints, and Procrustes-Aligned Mean Per Joint Position Error (PA-MPJPE) between the predictions and the ground-truth. All metrics are reported in $mm$.

The results in Table ~\ref{tab:comp-sota} show that our model in 3DPW reduces PA-MPJPE by 3.0mm, MPJPE by 3.1mm and MVE by 3.6mm when compared with BEDLAM-CLIFF (HRNet backbone). In EMDB reduces PA-MPJPE by 8.1mm, MPJPE by 11.6mm and MVE by 14.2mm when compared with BEDLAM-CLIFF (HRNet backbone). 

When compared with TokenHMR (ViT backbone) in 3DPW reduces PA-MPJPE by 0.4mm, MPJPE by 2.7mm and MVE by 4.3mm. In EMDB reduces MPJPE by 4.1mm and MVE by 7.2mm.

Furthermore, the results in Table~\ref{tab:comp-sota} demonstrate that training exclusively on synthetic data is effective and generalizes well to real-world data. %

In Table ~\ref{tab:comp_params} we compare the size of our model with that of the current state-of-the-art, using the number of parameters as a metric. Our method has 83.8\% less parameters that TokenHMR and 82\%   
less than HMR2.0. The reason that our model is significantly smaller is that we use a CNN backbone instead of ViT and also lightweight decoders and regressor.
\subsection{Qualitative Results}
Our qualitative results provide evidence on the effectiveness of our method across a diverse set of challenging scenarios. 
Figure~\ref{fig:qualitative2} consolidates results from four key datasets, illustrating the versatility and robustness of our approach across a variety of environments and challenges.

The top-left section of Figure~\ref{fig:qualitative2} showcases the 3D HPS estimation capabilities of D-PoSE on the 3DPW dataset, along with intermediate representations of depth and part segmentation. Despite the challenges posed by realistic outdoor scenes and occlusions, our method exhibits strong generalization, effectively transferring from synthetic training data to real-world environments. Its robustness is further evidenced by maintaining accuracy even in heavily occluded scenes, a common issue in real-world human pose estimation (HPS) applications.
Figure~\ref{fig:qualitative2} top-right presents results from the EMDB dataset, highlighting our method’s performance in a scene with a challenging pose. 
In the bottom-left of Figure~\ref{fig:qualitative2}, results from the synthetic BEDLAM dataset illustrate our method's ability to maintain high accuracy, validating its efficacy across both real and synthetic environments.
Finally, the bottom-right of Figure~\ref{fig:qualitative2} presents results from the RICH dataset, which features complex human poses.

\begin{table}[t]
\begin{tabular}{c|c|c|c}
\hline
\textbf{Dataset-Method} & \textbf{PA-MPJPE} & \textbf{MPJPE} & \textbf{MVE} \\ \hline
3DPW w/o Depth & 44.3 & \textbf{68.8} & \textbf{81.3} \\
3DPW with Depth & \textbf{43.6} & 68.9 & 81.4 \\ \hline
RICH w/o Depth & 50.1 & 80.6 & 92.1 \\
RICH with Depth & \textbf{47.8} & \textbf{77.0} & \textbf{87.8} \\ \hline
EMDB w/o Depth & 53.5 & 87.6 & 101.8 \\
EMDB with Depth & \textbf{53.2} & \textbf{85.5} & \textbf{99.0} \\ \hline
\end{tabular}
\caption{Ablation study on the impact of using (or not) depth. The results were obtained on the 3DPW, EMDB and RICH dataset.} 
\label{tab:abl_depth}
\end{table}

Figure~\ref{fig:header} shows some additional sample results obtained in images contained in the 3DPW and EMDB datasets.

Qualitative results also showcase the robustness of our method in diverse inputs regarding the race, gender and body-type of the person. These qualitative results underscore the generalization capability of our method and its potential to handle demanding real-world scenarios. 
\section{Ablation Study}
We conduct an ablation study on the impact of using depth in our model architecture. As shown in Table~\ref{tab:abl_depth}, the introduction of depth as an intermediate representation, in 3DPW improves PA-MPJPE  by 0.7mm. In the RICH dataset, MPJPE is reduced by 3.6mm, MVE by 4.3mm and PA-MPJPE by 2.3mm. In the EMDB dataset, MPJPE is reduced by 2.1mm, MVE by 2.8mm and PA-MPJPE by 0.3mm. We consider this a significant improvement in the challenging 3DPW, RICH and EMDB datasets.

\section{Conclusions}
We presented D-PoSE, a novel architecture for 3D human pose and shape estimation based on a single RGB frame. D-PoSE leverages depth as an intermediate representation, achieving state-of-the-art performance across all error metrics on the challenging 3DPW and EMDB datasets. Despite estimating both part-segmentation and depth maps, our approach significantly reduces the number of parameters compared to previous state-of-the-art methods. It trains in one stage, ensuring a straightforward and lightweight design that  makes it a strong foundation for future advancements in human pose estimation. Future work, could leverage temporal information from video input and/or incorporate larger transformer-based backbones such as Vision Transformers (ViT). 

\section*{Acknowledgments}
We thank Vassilis Nicodemou for his valuable assistance with the ablation study.
This work was co-funded by (a) the European Union (EU - HE Magician –
Grant Agreement 101120731) and, (b) the Hellenic Foundation for Research and Innovation (HFRI) under the ``1st Call for HFRI Research Projects to support Faculty members and Researchers and the procurement of high-cost research equipment'', project I.C.Humans, no 91. The authors also
gratefully acknowledge the support for this research from the VMware University Research Fund (VMURF).

{\small
\bibliographystyle{ieee_fullname}
\bibliography{egbib}
}

\end{document}